\documentclass[sigconf, nonacm]{acmart}

\AtBeginDocument{%
  \providecommand\BibTeX{{%
    \normalfont B\kern-0.5em{\scshape i\kern-0.25em b}\kern-0.8em\TeX}}}

\setcopyright{acmlicensed}
\copyrightyear{2018}
\acmYear{2018}
\acmDOI{XXXXXXX.XXXXXXX}

\acmConference[Conference acronym 'XX]{Make sure to enter the correct
  conference title from your rights confirmation emai}{June 03--05,
  2018}{Woodstock, NY}
\acmISBN{978-1-4503-XXXX-X/18/06}

\newcommand{\model}{TransFlower}

\usepackage{multirow}
\usepackage{graphicx}
\usepackage{subcaption}
\begin{document}

\title{\model: An Explainable Transformer-Based Model with Flow-to-Flow Attention for Commuting Flow Prediction}



\author{Yan Luo}
\affiliation{%
  \institution{Hong Kong Polytechnic University \& Massachusetts Institute of Technology}
  \country{China \& USA}
}
\email{yluo97@mit.edu}

\author{Zhuoyue Wan}
\affiliation{%
  \institution{Hong Kong Polytechnic University}
  \state{Hong Kong SAR}
  \country{China}
}
\email{zhuoywan@polyu.edu.hk}

\author{Yuzhong Chen}
\affiliation{%
  \institution{Arizona State University}
  \city{Tempe}
  \state{AZ}
  \country{USA}}
\email{cyz001383@gmail.com}

\author{Gengchen Mai}
\affiliation{%
  \institution{University of Georgia}
  \city{Athens}
  \state{GA}
  \country{USA}
}
\email{gengchen.mai25@uga.edu}

\author{Fu-lai Chung}
\affiliation{%
  \institution{Hong Kong Polytechnic University}
  \state{Hong Kong SAR}
  \country{China}
  }
\email{cskchung@polyu.edu.hk}

\author{Kent Larson}
\affiliation{%
  \institution{Massachusetts Institute of Technology}
  \city{Cambridge}
  \state{MA}
  \country{USA}}
\email{kll@media.mit.edu}

\renewcommand{\shortauthors}{Y. Luo, et al.}

\begin{abstract}

    Understanding the link between urban planning and commuting flows is crucial for guiding urban development and policymaking. This research, bridging computer science and urban studies, addresses the challenge of integrating these fields with their distinct focuses.
    Traditional urban studies methods, like the gravity and radiation models, often underperform in complex scenarios due to their limited handling of multiple variables and reliance on overly simplistic and unrealistic assumptions, such as spatial isotropy.
    While deep learning models offer improved accuracy, their ``black box'' nature poses a trade-off between performance and explainability — both vital for analyzing complex societal phenomena like commuting flows. 
    To address this, we introduce \model{}, an explainable, transformer-based model employing flow-to-flow attention to predict urban commuting patterns. It features a geospatial encoder with an anisotropy-aware relative location encoder for nuanced flow representation. Following this, the transformer-based flow predictor enhances this by leveraging attention mechanisms to efficiently capture flow interactions. Our model outperforms existing methods by up to 30.8\% Common Part of Commuters (CPC), offering insights into mobility dynamics crucial for urban planning and policy decisions. The code is accessible at: https://github.com/zwanah/TransFlower
\end{abstract}




\received{20 February 2007}
\received[revised]{12 March 2009}
\received[accepted]{5 June 2009}

\maketitle

\section{Introduction}

Cities, as the epicenters of economic, social, and cultural activities, play a pivotal role in the evolution of human civilization \cite{10.5555/2553103, bettencourt2007growth}. Understanding urban human mobility is crucial for creating equitable and sustainable cities, ensuring accessible transportation, economic opportunities, and enhanced quality of life for residents \cite{Gonzalez2008UnderstandingIH, Alessandretti2022WhatHM, barbosa2018human}. Commuting flows, defined as the movements between a worker's home and their place of employment, play a crucial role in revealing daily movement patterns within urban settings. These flows inform city planning, infrastructure development, and contribute to the understanding of urban dynamics and socioeconomic factors \cite{hincks2010spatial, sohn2005commuting}. For instance, some real-world questions we have encountered are, ``If a developer plans to introduce several corporate office buildings in a certain area, how would this affect the citywide commuting flows? How should the government appropriately respond by preparing the necessary supporting infrastructure?'' If we translate these questions into scientific terminology, they boil down to the problem of commuting flow prediction -- learning the correlation between urban planning and the distribution of commuting flows. There are two main practical applications. First, given a new urban development plan, it becomes feasible to determine the corresponding distribution of commuting flows, aiding in the assessment of its impact. Second, traditional methods of collecting commuting flow data rely on national statistical bureaus \cite{mckenzie2013county}, which are costly and untimely. With a reliable prediction model, we can achieve real-time estimation of commuting flows.

The task of commuting flow prediction sits at the crossroads of urban studies and computer science, each bringing distinct approaches and limitations. Traditionally, urban studies have depended on physical models like the gravity model \cite{stewart1941inverse, zipf1946p, erlander1990gravity, barthelemy2011spatial} and radiation model \cite{simini2012universal} to predict and analyze urban flows, primarily due to limitations in data availability and quantitative analysis capabilities. These models offer a theoretical framework for understanding urban dynamics. However, these models face significant limitations, such as simple assumptions and limited performances \cite{barbosa2018human, simini2012universal}. The demand for enhanced performance in this task is crucial, as it allows stakeholders to assess the feasibility of specific planning initiatives and interventions with greater performance. This requirement for performance necessitates access to more comprehensive datasets and a shift away from traditional models towards more sophisticated methodologies. In response to these challenges, deep learning models from computer science have emerged as a powerful alternative \cite {Liu2020LearningGE, Zhou2023TowardsGM, 10.1145/3511808.3557255}, offering the promise of higher performance by leveraging large volumes of data to learn complex patterns of urban mobility. Despite their success in achieving improved predictive performance, these models often suffer from a lack of explainability or provide insights that are too shallow for practical application in urban planning. In the context of social phenomena like commuting flows, the importance of explainability is on par with performance. A thorough understanding of the underlying forces and dynamics of mobility is crucial for validating the model's credibility and supporting planning and policy development efforts. This opacity poses a dilemma, as there traditionally exists a trade-off between deep learning models' performance and explainability \cite{DARPA2016XAI}. Recent advances within the Explainable Artificial Intelligence (XAI) community, however, demonstrate that it is possible to design models that do not sacrifice model explainability for performance \cite{rudin2019stop}. Some works point out that creating an explainable architecture involves designing with explainability in mind, such as utilizing fewer hidden layers or learnable parameters or adopting a modular design, so that every part of the model has a clear function and explanation \cite{friston2024designing, 8466590}. Inspired by this development, an intriguing research question emerges: \textit{"How to design a model for commuting flow prediction that achieves both high performance and good explainability?"}

To answer this question, we conducted an in-depth analysis of the two classical models previously mentioned: the gravity model and the radiation model. The gravity model predicts flow distribution by mimicking the gravitational forces observed in the physical world, suggesting that the volume of commuting flow is directly proportional to the attraction between two locations and inversely proportional to the distance between them. Despite its explanatory power through a clear, parameter-based formula, the gravity model's reliance on region-specific adjustable parameters and known analytical inconsistencies limits its applicability \cite {simini2012universal}. In response, the radiation model \cite{simini2012universal} was developed to overcome many of the gravity model's drawbacks, using a stochastic process to simulate local mobility decisions and analytically deriving commuting and mobility fluxes based on population distribution alone. Nonetheless, the radiation model's reliance on a limited set of variables (primarily population and distance) results in low performance due to the omission of critical geographical details like land use types and points of interest (POIs). 
Moreover, it assumes a isotropic ideal space by simple distance decay for flow determinants, disregarding the diverse factors and complex spatial correlations influencing human mobility, such as transportation conditions, topography, and land use patterns, which leads to inaccuracies in representing the actual structure and variability of flows \cite{barbosa2018human}. The limitations of the former can be addressed by integrating comprehensive geographic data of urban regions into deep-learning models, while for the latter, we aim to develop a module within our model capable of adaptively learning location characteristics to circumvent the unrealistic assumption of spatial isotropicity \cite{mai2018adcn,zhu2019making,zhang2009conserved}.

Additionally, it is worth noting that among the methods involving deep learning, one particular study, DeepGravity \cite{simini2021deep}, touches upon explainability. DeepGravity claims to provide explainability through SHAP \cite{lundberg2017unified}, a game theory-based approach for interpreting machine learning models. However, such model-agnostic explanation methods can only superficially mimic the input-output relationship of the actual model, without the ability to deeply analyze causal relationships. A fundamental reason for the lack of deep explainability for DeepGravity is the oversimplified modeling that relies solely on feed-forward networks to encode information from individual flows, thereby overlooking the interplay between flows. Given that commuting flows represent a complex network challenge, understanding the intricate dynamics and interplay between flows is crucial. These interactions are key to deciphering the mechanisms of information transmission within the network and understanding how the network's structure impacts these processes \cite{donner2017introduction}. Therefore, incorporating interactions between flows could not only enhance the model's explainability but also potentially contribute to an improvement in performance.

In light of these considerations, we have approached the design of our solution from first principles, resulting in the creation of a minimalist yet effective model. We introduce \model{}, an explainable transformer-based \cite{vaswani2017attention} model featuring flow-to-flow attention mechanisms specifically tailored for the predicting of commuting flows. The transformer is selected primarily for its efficiency in handling the interactions among items within a sequence or a set, enabling intricate flow relationships to be modeled. Additionally, the attention mechanism enhances model explainability. Specifically, we first employ a novel relative location encoder to capture the spatial relationship between the origin and destination regions of individual flows. While the DeepGravity model only considers the distance between origins and destinations, with the help of relative location encoding, our model can jointly consider both distance and direction information. 
This encoder is designed to adaptively learn spatial representations, effectively countering the radiation model's unrealistic assumption of spatial uniformity and isotropicity. 
Subsequently, we integrate the spatial relationship information with other geographic features of the flow's origin and destination regions in the geo-spatial encoder. This integrated data is then fed into a transformer-based flow predictor designed to learn the interactions between different flows. Finally, we utilize attention mechanisms between flows to investigate the relationship between commuting flows and urban features, offering new insights into the dynamics of urban mobility.

Our experiments, conducted across three states in the US, indicate that our framework outperforms existing state-of-the-art deep learning models by up to 30.8\% in terms of the metric Common Part of Commuters (CPC). Additionally, we illustrate how the flows generated by our model can be explained through flow-to-flow attention mechanisms. Moreover, we prove that our relative location encoder is capable of capturing spatial anisotropicity. With our model, stakeholders will be equipped to precisely describe the dynamics of commuting flows, a crucial aspect in shaping cities in alignment with SDGs.

\section{Problem Formulation}
It is noteworthy that the prediction of commuting flows differs from the more common spatiotemporal traffic flow forecasting problem \cite{lana2018road,  vlahogianni2014short}. Traffic flow is a short-term, dynamic, and time-series-related phenomenon, whereas commuting flows represent long-term, relatively static, and time-series-independent repeated daily movements. In traffic flow modeling, historical flows are used as input features, whereas the inputs for commuting flow prediction problems are the geographical attributes of urban regions.

\textit{Definition 1} \textbf{Urban Region}: Urban regions $u_i$ can be census blocks, block groups, census tracts, counties, or states. To ensure our model has an appropriate number of samples in each experimental area, in this paper, we use block groups as urban regions. Each region is characterized by a variety of geo-spatial features, including place features (e.g., POIs within the region) and location information. Place features of each region can be represented as $x_{u_i}$. The centroid coordinates of each region can be indicated by $loc_{u_i}$. The distance measured along the surface of the earth between the centroid of region $u_i$ and region $u_j$ is denoted as $r_{ij}$.

\textit{Definition 2} \textbf{Commuting Flow}: Commuting flows are a set of 3-tuple $f_{ij}=(o_i, d_j, v_{ij})$ where $o_i$ is the flow origin urban region, $d_j$ is the flow destination urban region, $v_{ij}$ is the number of commuters moving from $o_i$ to $d_j$ per unit time. 
We use $S = \{f_{kl}=(o_k, d_l, v_{kl})\}$ to denote the full set of all urban commuting flows and 
$S_i = \{f_{il}=(o_i, d_l, v_{il})\}  \subseteq S$ be the subset of $S$ denoting all flows originating from region $o_i$.
Then the total outflow, $O_i$, from region $o_i$ is the total number of flows per unit time originating from region $o_i$, i.e., $O_i=\sum_{f_{il} \in S_i} v_{il}$.

\textit{Problem Statement} \textbf{Commuting Flow Prediction}: Essentially, this can be considered a task of outflow allocation. We can first forecast the probabilities of destinations for flows originating from a specific origin region. By multiplying these predicted probabilities with the known outflows, we can determine the actual flows.

\section{Methodology}
\begin{figure*}[t]
  \centering
  \includegraphics[width=0.9\textwidth]{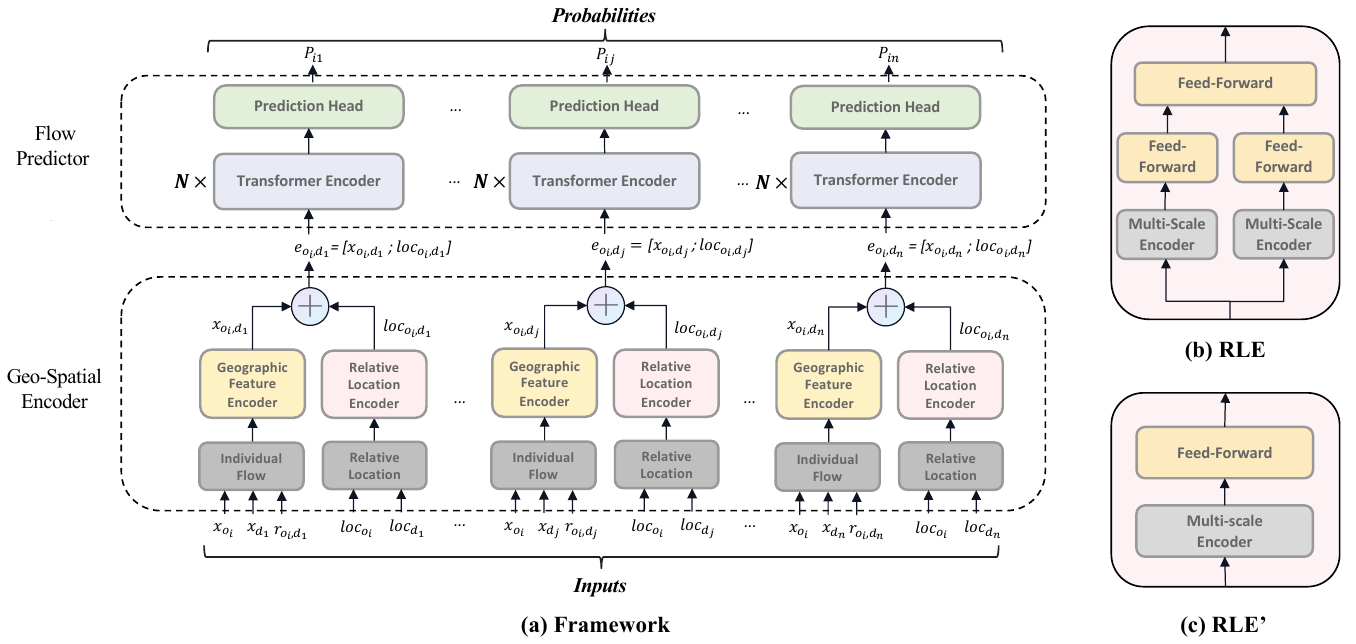}
  \caption{The framework of \model{} which contains two major components: geo-spatial encoder and flower predictor.}
  \label{fig:framework}
\end{figure*}

Considering the geo-spatial features $x_{u_i}$ and $loc_{u_i}$ of urban regions, the output of \model{} is an $n$-dimensional vector $P_{i,j}$, where $j=1,...,n$, representing the probability of flows originating from $o_i$ and arriving at each destination $d_j$. As illustrated in Figure \ref{fig:framework}, the model employs two major components to achieve this: the geo-spatial encoder and the flow predictor.

\subsection{Geo-Spatial Encoder}
Predicting commuting flows fundamentally involves geo-spatial information.
To this end, we introduce the Geo-Spatial Encoder. The input to the Geo-Spatial Encoder is divided into three main categories: the place features $x_{u_i}$ of both the origin and destination regions, the coordinate information $loc_{u_i}$ of these regions, and the distance $r_{o_i, d_j}$ between them. 
Following the practice of DeepGravity \cite{simini2021deep}, we concatenate the place features of the origin and destination, i.e., $x_{o_i}$ and $x_{d_j}$, as well as the distance $r_{o_i, d_j}$, and feed these features to a feed-forward neural network to obtain an embedding capturing the characteristics of both $o_i$ and $d_j$.

However, merely accounting for the distance $r_{o_i, d_j}$ falls short of capturing comprehensive spatial information, as commuting flows exhibit anisotropic patterns \cite{zhang2009conserved}, indicating that direction is also a crucial piece of information to consider.
Thus, we compute the relative location ${rl}_{ij} = loc_{o_i} - loc_{d_j}$ ($\in \mathbb{R}^2$) and feed it into a specially designed relative location encoder based on Space2Vec \cite{Mai2020Multi-Scale}, to jointly capture the distance and direction information.

\subsubsection{Geographic Feature Encoder}

To extract information on individual flows, we follow the practice of benchmark model DeepGravity, by using place features and the distance feature as inputs for our flow geographic feature encoder.

Place features, denoted by $x_{u_i}$ ($u_i$ can be either $o_i$ or $d_j$), delineate various geographical aspects of areas, such as the number of retail establishments present within a specific region. The dimensionality of these place features corresponds to the aggregate count of features evaluated. Among the place features utilized are the population magnitude of each region and 19 attributes derived from OpenStreetMap \cite{OpenStreetMap2017}, spanning the following categories:

\begin{itemize}
\item Facilities related to food (4 attributes: point-amenity, point-shop, polygon-amenity, polygon-shop): the comprehensive tally of POIs and structures associated with any food-related service, e.g., restaurants, fast food outlets, and bars.
\item Facilities related to retail (4 attributes: point-amenity, point-shop, polygon-amenity, polygon-shop): the comprehensive tally of POIs and structures associated with retail services, e.g., marketplaces, pharmacies, and convenience stores.
\item Facilities related to education (3 attributes: point-amenity, polygon-amenity, polygon-building): the comprehensive tally of POIs and structures associated with educational institutions, e.g., schools, colleges, and universities.
\item Facilities related to health (3 attributes: point-amenity, polygon-amenity, polygon-building): the comprehensive tally of POIs and structures associated with health services, e.g., dentists, hospitals, and clinics.
\item Facilities related to transport (5 attributes: point-amenity, point-public-transport, polygon-amenity, polygon-building, polygon-public-transport): the comprehensive tally of POIs and structures associated with transport services, e.g., parking lots, bus stations, and charging stations.
\end{itemize}

We concatenate the distance feature $r_{i,j}$ with a 40-dimensional vector of place features, $x_{u_i}$, which includes 19 geographic attributes from both the origin and destination, in addition to their respective population counts, to formulate the 41-dimensional feature embedding that define individual flow patterns. These consolidated features are then subjected to a transformation through a single-layer feed-forward network, resulting in a 256-dimensional individual flow representation $x_{o_i, d_j}$. This process effectively projects the lower-dimensional features into a higher-dimensional space.

\subsubsection{Relative Location Encoder}
To overcome the unrealistic assumption of spatial isotropicity in classical models, we introduce a relative location encoder to adaptively learn the representation of spatial information for each flow. While there has been some work on spatial representation learning \cite {10.1145/3511808.3557720, 10.1145/3583780.3615083}, most of it deals with specific tasks and often disregards geographic coordinates. Space2Vec \cite{Mai2020Multi-Scale} is a general-purpose representation learning model that encodes the absolute positions or spatial relationships of places into a vector space representation. This work is inspired by the Nobel Prize-winning neuroscience research that demonstrated mammalian grid cells provide a multi-scale periodic representation \cite{Abbott2014NobelPrize}. It has been discovered that leveraging location embeddings, generated through coordinate decomposition, assists deep-learning models in overcoming the limitations inherent in traditional single-scale approaches by adopting multi-scale representations \cite{mac2019presence}.

Inspired by the multi-scale representations of Space2Vec, we develop a relative location encoder that comprises two key components: a multi-scale encoder and a feed-forward network. Unlike Space2Vec, which operates on single coordinate points, our focus is on flow. We calculate the relative location ${rl}_{ij} = loc_{o_i} - loc_{d_j}$ 
as the inputs for our relative location encoder. 

More precisely, our relative location encoder represents the relative location ${rl}_{ij}$ in a multi-scale relative location embedding. While traditional transformer position encoding layers operate in discrete 1-dimensional spaces, such as word positions in a sentence, we tackle higher-dimensional continuous spaces, like the Earth's surface. To encompass a 2-dimensional space, we utilize a set of sine and cosine functions at various frequencies, amounting to $freq$ in total, to encode the relative location ${rl}_{ij}$. The multi-scale encoder is denoted as $Enc\left({rl}_{ij}\right)=\mathbf{NN}\left(PE({rl}_{ij})\right)$, where $\mathbf{NN}$ consists of fully connected ReLU layers. Here, $PE({rl}_{ij}) = \left[PE_0({rl}_{ij}); \ldots ; PE_{K-1}({rl}_{ij})\right]$ concatenates multi-scale representations. $K$ is the total number of scales.

We decompose and project the relative location ${rl}_{ij}$ into the direction of three base vectors in $\mathbb{R}^2$, named $a_1$, $a_2$, $a_3$, each distinctly angled at $2\pi/3$ from one another. According to Space2Vec, a typical configuration might set $\mathbf{a}_1 = [1,0]^T$, $\mathbf{a}_2 = [-1/2,\sqrt{3}/2]^T$, and $\mathbf{a}_3 = [-1/2,-\sqrt{3}/2]^T$, establishing a symmetrical geometric structure for spatial encoding. $\lambda_{min}$ and $\lambda_{max}$ represent the minimum and maximum scales, respectively, with $g = \frac{\lambda_{max}}{\lambda_{min}}$. For each scale $s$, $PE_s({rl}_{ij})$ combines three components as follows:
\begin{align}
PE_{s, j}({rl}_{ij})=\left[\cos \left(\frac{\left\langle{rl}_{ij}, \mathbf{a}_j\right\rangle}{\lambda_{\min } \cdot g^{s /(S-1)}}\right) ; \sin \left(\frac{\left\langle{rl}_{ij}, \mathbf{a}_j\right\rangle}{\lambda_{\min } \cdot g^{s /(S-1)}}\right)\right], \forall j=1,2,3 ;
\label{equ:petheory}
\end{align}
where $\langle {rl}_{ij},a_j \rangle$ is the inner product of ${rl}_{ij}$ and $a_j$.

It's noteworthy that an analysis of Figure \ref{fig:sub4} revealed intriguing experimental phenomena (to be detailed in Section \ref{sec:residual}), prompting a reevaluation of the design of the relative location encoder. We proposed two designs for the Relative Location Encoder (RLE):

\begin{itemize}
    \item \textbf{RLE}: As shown in Figure \ref{fig:framework}b, the encoder includes two branches, each with a multi-scale encoder and a feed-forward network. Two encoders have distinct base vector sets. An additional feed-forward layer merges outputs from both branches, creating $loc_{o_i, d_j}$ which captures the complex geographical relationships between origin and destination regions at multiple levels.
    \item \textbf{RLE'}: Illustrated in Figure \ref{fig:framework}c, this encoder variant operates with a single branch, utilizing the typical configuration of base vectors' orientations as mentioned above. The output from this branch, $loc_{o_i, d_j}$, is the relative location vector.
\end{itemize}

Subsequently, by concatenating the individual flow embedding $x_{o_i, d_j}$ with the flow spatial embedding $loc_{o_i, d_j}$ generated by the relative location encoder, we create a comprehensive embedding for each flow, represented as $e_{o_i, d_j} = \left[x_{o_i, d_j}; loc_{o_i, d_j}\right]$. This process effectively combines specific flow characteristics with spatial information, facilitating a more nuanced analysis of commuting patterns.

\subsection{Flow Predictor}
In the flow predictor, the generated flow embeddings from the geo-spatial encoder are processed through a $N$-layer transformer encoder module to 
model the interactions among different flows that share the same origin with the self-attention mechanism. This is followed by a prediction head, which includes a feed-forward layer and a softmax layer, to generate the probability distribution of flows from a specified origin region to various destination regions. This structure ensures that the model captures the complex dynamics and interactions between different commuting flows, enhancing its predictive performance and relevance for urban planning and analysis.

\subsubsection{Transformer Encoder}

Our choice of a transformer-based model is grounded in its sequential processing capabilities, allowing for effective representation of flow relationships and complex mobility patterns. Its attention mechanism offers explicit input prioritization and improved explainability through visual insights into predictive focuses. Furthermore, transformers' contextual processing and parallelism enhance efficiency, making them ideal for large-scale and complex commuting flow analyses.

Specifically, each transformer encoder layer consists of a multi-head self-attention module and a feed-forward network. Residual connections and layer normalizations are also employed. In each attention head, self-attention for flow $f_{ij}$ can be formulated as:

\begin{equation}
e_i \prime = e_i + w_z \sum_{j=1}^n \frac{\exp(w_q e_i \times w_k e_j)}{\sum_{m=1}^n \exp(w_q e_i \times w_k e_m)} w_v e_j
\end{equation}
where $e_i$ and $e_i\prime$ are the input and output embeddings, and $w_{\{q, k, v, z\}}$ denotes linear transform weights for the query, key, value, and output matrices.

\subsubsection{Prediction Head}

The outputs from transformer encoder layers are then fed into a feed-forward layer, which leads to a score $s(o_i, d_j)$ ranging from $-\infty$ to $+\infty$. For any region pair $(o_i, d_j)$, a higher score indicates a greater likelihood of observing a flow from $o_i$ to $d_j$. Subsequently, a softmax function converts these scores into probabilities $P_{i,j}$, ensuring they are positive and collectively sum to one. The actual flow between two regions is then determined by multiplying this probability with the total outflow from the origin region.

\subsubsection{Loss Function}
The model's loss function is defined as cross-entropy, as it measures the difference between the predicted probabilities and the actual distribution, effectively guiding the model to improve its predictions over time. The loss can be denoted as:
\begin{equation}
H=-\sum_i \sum_j \frac{f_{ij}\left(o_i, d_j\right)}{O_i} \ln P_{i, j}
\end{equation}
where $\frac{f_{ij}\left(o_i, d_j\right)}{O_i}$ represents the proportion of observed flows departing from origin $o_i$ to destination $d_j$, while $P_{i, j}$ signifies the predicted probability by the model for a flow from $o_i$ to $d_j$. The aggregation of cross-entropy values across different origins, denoted by the sum over $i$, is based on the premise that flows originating from disparate locations are considered independent events. This assumption enables the utilization of cross-entropy's additive nature for independent variables, facilitating a comprehensive evaluation of the model's performance across multiple origins.

\section{Results}

In this section, we conduct extensive experiments on three datasets to evaluate our proposed model.

\subsection{Experiment Setups}
\subsubsection{Datasets}
We evaluated our proposed model on real-world commuting flow datasets collected from California, Massachusetts, and Texas. Our input data for urban regions was sourced from the 2019 American block group’s cartographic boundary files provided by the US Census Bureau. Due to computational considerations, we selectively focused on specific areas within California (7872 block groups), Massachusetts (2350 block groups), and Texas (7900 block groups), ensuring coverage of the most central and significant locations. For instance, the regions selected in California include key areas such as Los Angeles and San Francisco, highlighting our model's applicability to diverse and densely populated urban environments.

For commuting flows, we utilized the 2019 Origin-Destination Employment Statistics (LODES) dataset \cite{USCensusBureau2019LEHD}. This dataset is updated annually and captures the home and employment locations of workers, thus offering insights into stable commuting patterns. The commuting flows are then aggregated to the geographic unit level for analysis. In our experimental area in California, a total of 2,552,243 origin-destination pairs accounted for 5,583,372 commuting flows. The Massachusetts dataset included 58,235 origin-destination pairs, leading to 1,404,860 commuting flows. In Texas, there were 756,179 origin-destination pairs, resulting in 1,688,550 commuting flows.

For place attributes, we employ point and polygon data from OpenStreetMap \cite{OpenStreetMap2017}, a widely recognized platform for volunteer-contributed geographic information. Each point and polygon represents a geographic feature with a specific purpose. Typically, a polygon is used to delineate the outline of a building. For example, Union Station could be depicted as either a point or a polygon, categorized under public transportation. Conversely, a restaurant located within Union Station would be represented as an independent point separate from Union Station, and it would fall under the category of dining.

\subsubsection{Evaluation Metrics}
To evaluate the performance of our predictions, we utilize three widely recognized metrics: Common Part of Commuters (CPC), Mean Absolute Error (MAE), and Root Mean Squared Error (RMSE) \cite{barbosa2018human, simini2021deep}. MAE and RMSE offer a straightforward and effective means of assessing the magnitude of errors in regression analyses. CPC quantifies the similarity between the predicted flow volume $v^p\left(o_i, d_j\right)$ and real flow volume $v^r\left(o_i, d_j\right)$, which can be denoted as:
\begin{equation}
CPC=\frac{2\sum_{i, j} \min \left(v^p\left(o_i, d_j\right), v^r\left(o_i, d_j\right)\right)}{\sum_{i, j} v^p\left(o_i, d_j\right)+\sum_{i, j} v^r\left(o_i, d_j\right)}.
\end{equation}

\subsubsection{Implementations}

To align with the settings used in DeepGravity, we randomly partition the commuting flows into training and validation datasets using an 80:20 split. The network training employs the early stopping technique to prevent overfitting, with a patience setting of 20 epochs and utilizes the RMSprop optimizer. This optimizer is configured with a momentum of 0.9 and a learning rate of 0.0001, processing batches containing 512 origin regions each. The feed-forward network features a hidden size of 256, and we set the network's random seed to 1234. A dropout rate of 0.1 is applied to mitigate overfitting, and the Transformer encoder is structured with two layers.

To streamline the training process of the transformer and ensure alignment with DeepGravity's methodology, we select 256 destination regions for each origin region. This number approximates the average count of flows originating from each region. If the actual number of destinations surpasses 256, we employ random sampling to select 256 destinations; Otherwise, we utilize zero-padding to achieve this count.

Regarding the parameters within the relative location encoder, the maximum lambda $\lambda_{max}$ is set to the maximum diameter of the study area, which is 20013 m for California, 20009 m for Massachusetts, and 20013 m for Texas. 
The number of different sinusoidal with different frequencies $freq$ and minimum scale $\lambda_{min}$ are hyperparameters, optimized differently across study areas. For instance, for the California dataset, $freq$ is set to 16, and the $\lambda_{min}$ to 1.

\subsubsection{Baselines}
To demonstrate our model's efficacy, we conduct comparisons against the following baseline models:
\begin{itemize}
\item Gravity \cite{barbosa2018human}: This adaptation of the classic gravity model incorporates two power exponents, suggesting that the flow of people or goods between two areas directly correlates with their propulsiveness (outflow) and attractiveness (inflow), while being inversely related to the distance separating them. The model's inputs exclude geographic features.
\item Radiation \cite{simini2012universal}: The radiation model posits that urban mobility is determined by the city's population density distribution. The likelihood of movement from origin $o_i$ to destination $d_j$ is denoted by the following formulation:
\begin{equation}
P_{i,j}=\frac{p_i p_j}{\left(p_i+S_{i,j}\right)\left(p_i+p_j+S_{i,j}\right)}
\end{equation}
where $p_i$ represents the population of the origin region $o_i$, and $S_{i,j}$ denotes the total population within the circle centered at $o_i$ with a radius equal to the distance between the origin and destination. The model's inputs exclude geographic features.
\item Random Forest (RF) \cite{Pourebrahim2019TripDistribution}: As a classical 
machine learning method \cite{breiman2001random}, RF is reported as a strong baseline 
\cite{Pourebrahim2019TripDistribution}. The inputs 
are consistent with ours.
\item GMEL \cite{Liu2020LearningGE}: This model forms an adjacency graph and utilizes two attention-based graph neural networks (GATs) to independently learn the embeddings for origins and destinations. Additionally, the authors designed two supplementary tasks for forecasting inflow and outflow. To make a fair comparison with our model, we did not adopt this supplementary part. 
The inputs 
are consistent with ours.
\item DeepGravity \cite{simini2021deep}: This model views flow prediction as a task of allocating outflow and forecasts the probabilities of destinations for flows originating from a specific origin region. The inputs 
are consistent with ours.
\end{itemize}

\subsection{Performance Analysis}

We evaluated the performance of both baseline models and our proposed model using a test set, with detailed results presented in Table \ref{tab:perform}. \model{} stands out as the top performer across all datasets and metrics, achieving up to a 30.8\% improvement on CPC, thereby affirming the effectiveness of our approach. Our analysis offers several key insights.

Among the models assessed, the Radiation model ranks lowest in performance. This shortfall is likely due to its oversimplified assumptions like spatial anisotropy. Conversely, models incorporating geographic features (i.e., RF, GMEL, DeepGravity, \model{}) markedly surpass those lacking these elements (i.e., Gravity and Radiation). This finding emphasizes the critical role of geographic features in the accurate prediction of commuting flows. By embedding spatial contextual information, these models gain a nuanced understanding and depiction of commuting flow dynamics, resulting in enhanced performance across a variety of test scenarios.

\model{}'s superiority over the Gravity model demonstrates how a relative location encoder, capable of assimilating spatial anisotropy, combined with a sophisticated neural network framework, can significantly improve commuting flow prediction performance. Its advantage over DeepGravity further accentuates the value of capturing flow interactions via transformer encoder layers. Additionally, \model{} outperforms GMEL, a model leveraging GAT, indicating that transformers' implicit approach to modeling flow relationships is more efficacious than GAT's explicit method based on geographical proximity. This suggests that in the complex network challenge of commuting flows, the interplay between flows is influenced not solely by geographical proximity but also by complex, latent higher-order relationships.

\begin{table}[t!]
\caption{Performance analysis}
\label{tab:perform}
\resizebox{\columnwidth}{!}{%
\begin{tabular}{ccccc}
\hline
\textbf{Dataset} & \textbf{Model Name} & \textbf{CPC} & \textbf{MAE} & \textbf{RMSE} \\ \hline
\multirow{6}{*}{California}& Gravity& 0.6847& 1.1626& 3.5443\\
 & Radiation& 0.4074& 2.0782&6.4025\\
 & RF          & 0.6451          & 1.2987          &3.2309          \\
                           & DeepGravity & 0.6571          & 1.2475          & 3.1099          \\
                           & GMEL        & 0.6533          & 1.2563          & 3.1407          \\
                           & \model{} & \textbf{0.7691} & \textbf{0.8380} & \textbf{2.0780} \\ \hline
\multirow{6}{*}{Massachusetts}& Gravity& 0.6403& 1.5147& 3.5032\\
 & Radiation& 0.3221& 2.6807&6.2174\\
 & RF          & 0.6829          & 1.4343          &3.0398          \\
                           & DeepGravity & 0.6871          & 1.4151          & 2.9036          \\
                           & GMEL        & 0.6831          & 1.4347          & 2.9233          \\
                           & \model{} & \textbf{0.7573} & \textbf{1.0980} & \textbf{2.2700} \\ \hline
\multirow{6}{*}{Texas}& Gravity& 0.5886& 1.7434& 6.1830\\
 & Radiation& 0.3060& 2.6764&7.2387\\
 & RF          & 0.4811          & 0.6517          &3.0475          \\
                           & DeepGravity & 0.4786          & 0.6819          & 3.0313          \\
                           & GMEL        & 0.4535          & 0.6942          & 3.0753          \\
                           & \model{} & \textbf{0.6262} & \textbf{0.4794} & \textbf{2.3590} \\ \hline
\end{tabular}%
}
\end{table}

\subsection{Ablation Study}
As shown in Table~\ref{tab:ablation}, we validate the effectiveness of the key components contributing to performance enhancement: the transformer module and the relative location encoder (RLE). We tested three variants of \model{} to confirm this: (1) without RLE, (2) with RLE', and (3) without the transformer module (i.e., DeepGravity with RLE). The experimental results indicate that removing the RLE leads to a minimum drop of 9.7\% in CPC. 

\begin{table}[ht]
\caption{Ablation Study}
\label{tab:ablation}
\resizebox{\columnwidth}{!}{%
\begin{tabular}{ccccc}
\hline
\textbf{Dataset} & \textbf{Model Name} & \textbf{CPC} & \textbf{MAE} & \textbf{RMSE} \\ \hline
\multirow{4}{*}{California} & DeepGravity w/ RLE & 0.7337          & 0.9779          & 2.4580          \\
         & w/o RLE & 0.6820          & 1.1491          & 2.8219          \\
                           & w/ RLE' & 0.7602          & 0.8708          & 2.1120          \\
                           & w/ RLE (i.e., \model{}) & \textbf{0.7691} & \textbf{0.8380} & \textbf{2.0780} \\ \hline
\multirow{4}{*}{Massachusetts}& DeepGravity w/ RLE & 0.7333& 1.2060& 2.3950\\
				 & w/o RLE & 0.6905          & 1.3997          & 2.8613          \\
                           & w/ RLE' & 0.7499          & 1.1310          & 2.3190          \\
                           & w/ RLE (i.e., \model{})  & \textbf{0.7573} & \textbf{1.0980} & \textbf{2.2700} \\ \hline
\multirow{4}{*}{Texas} & DeepGravity w/ RLE & 0.5837& 0.5339& 2.5580\\
				 & w/o RLE & 0.4990          & 0.6265          & 2.8168          \\
                           & w/ RLE' & 0.6104          & 0.4959          & 2.4440          \\
                           & w/ RLE (i.e., \model{})  & \textbf{0.6262} & \textbf{0.4794} & \textbf{2.3590} \\ \hline
\end{tabular}%
}
\end{table}

\subsection{Residual Analysis}
\label{sec:residual}
\begin{figure*}[t!]
    \centering
    \begin{subfigure}[b]{0.33\textwidth}
        \centering
        \includegraphics[width=\textwidth]{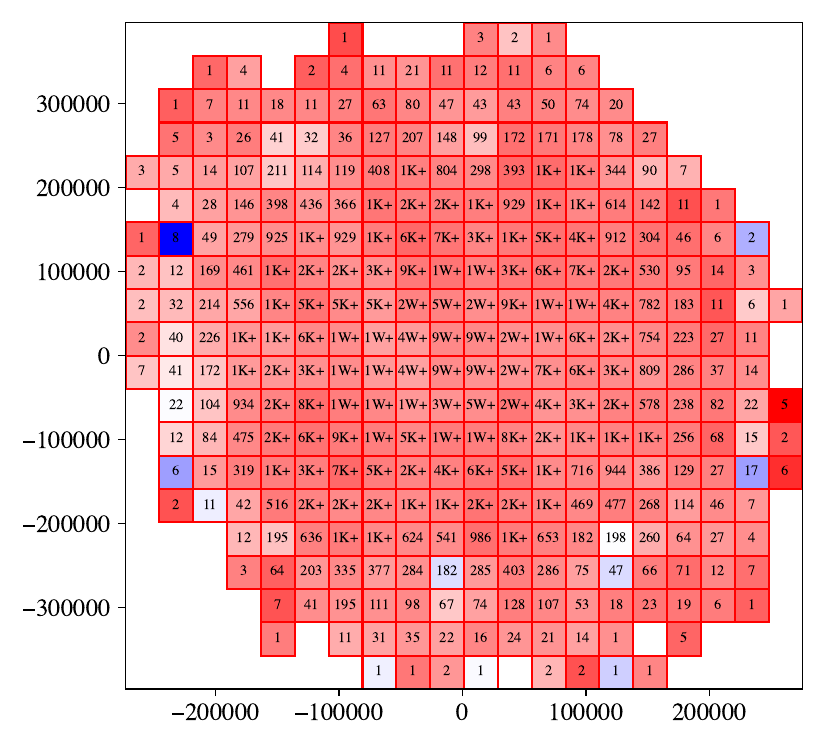}
        \caption{w/ RLE - DeepGravity}
        \label{fig:sub1}
    \end{subfigure}
    \hfill 
    \begin{subfigure}[b]{0.33\textwidth}
        \centering
        \includegraphics[width=\textwidth]{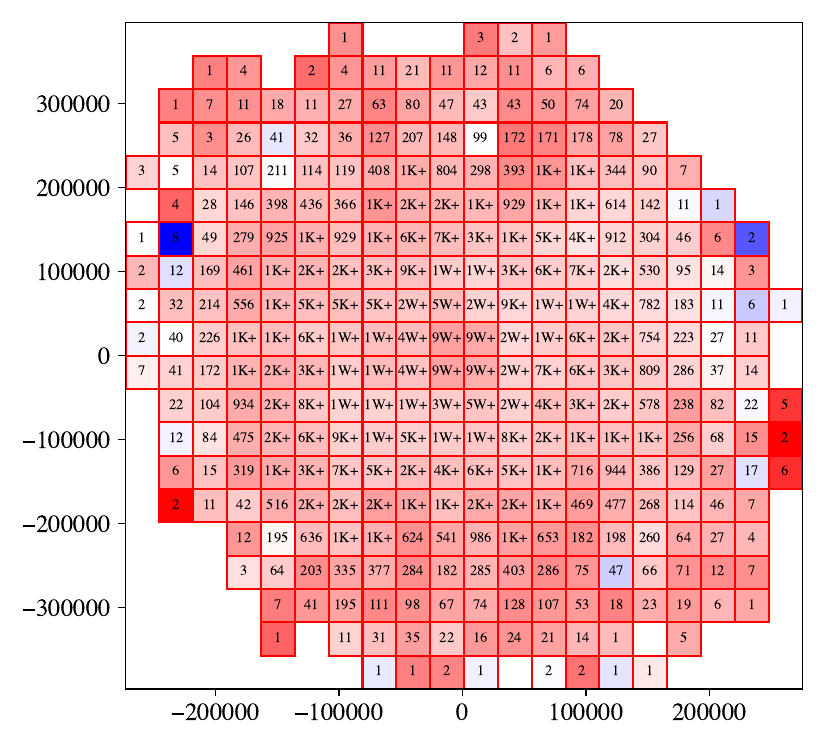}
        \caption{w/ RLE - DeepGravity w/ RLE}
        \label{fig:sub2}
    \end{subfigure}
    \hfill 
    \begin{subfigure}[b]{0.33\textwidth}
        \centering
        \includegraphics[width=\textwidth]{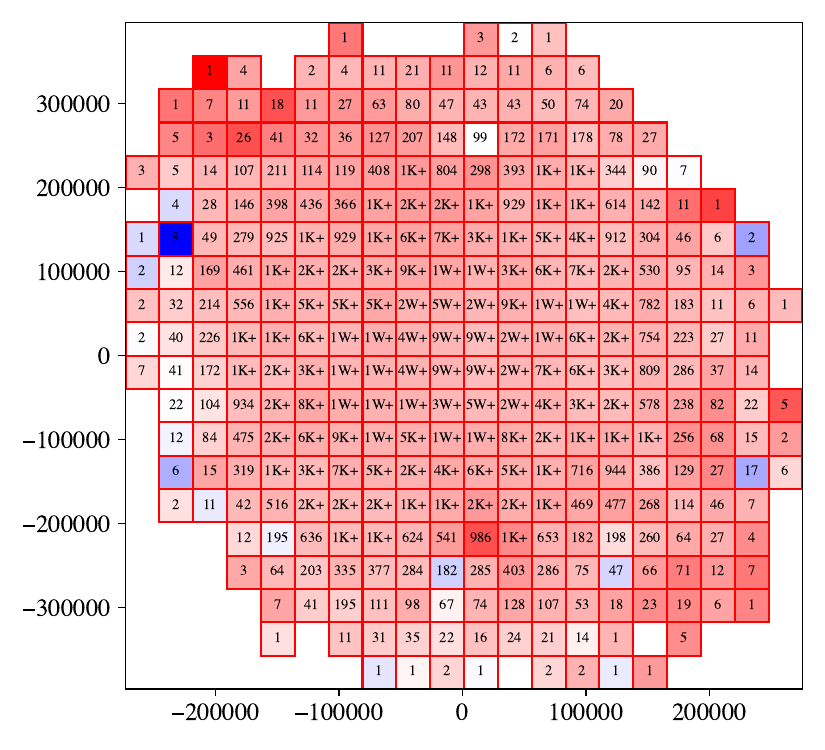}
        \caption{w/ RLE - w/o RLE}
        \label{fig:sub3}
    \end{subfigure}
    \vspace{1em}
    \begin{subfigure}[b]{0.33\textwidth} 
        \centering
        \includegraphics[width=\textwidth]{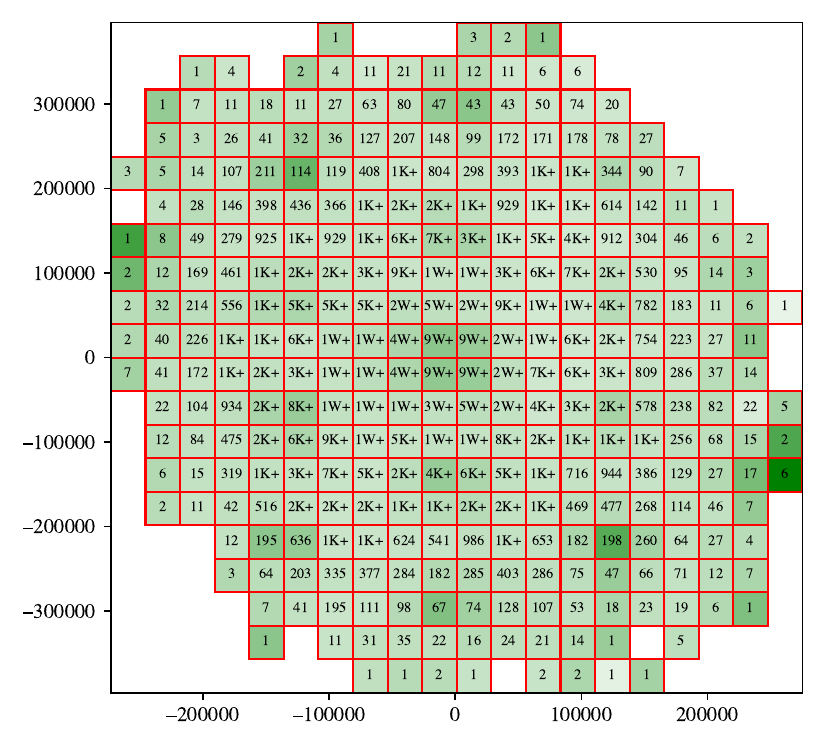}
        \caption{w/ RLE'} 
        \label{fig:sub4}
    \end{subfigure}
    \begin{subfigure}[b]{0.33\textwidth}
        \centering
        \includegraphics[width=\textwidth]{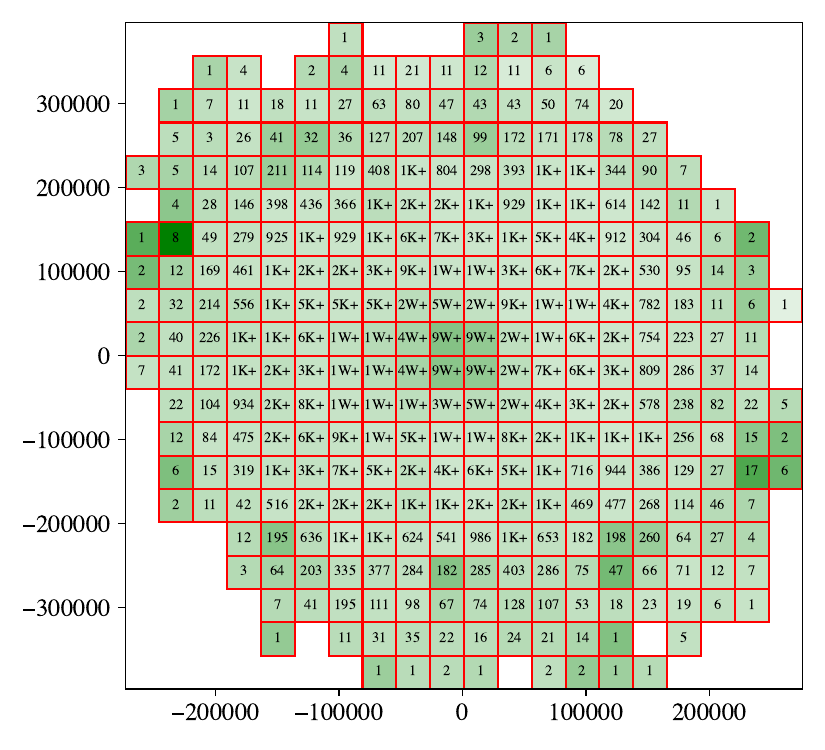}
        \caption{w/ RLE}
        \label{fig:sub5}
    \end{subfigure}
    \begin{subfigure}[b]{0.33\textwidth}
        \centering
        \includegraphics[width=\textwidth]{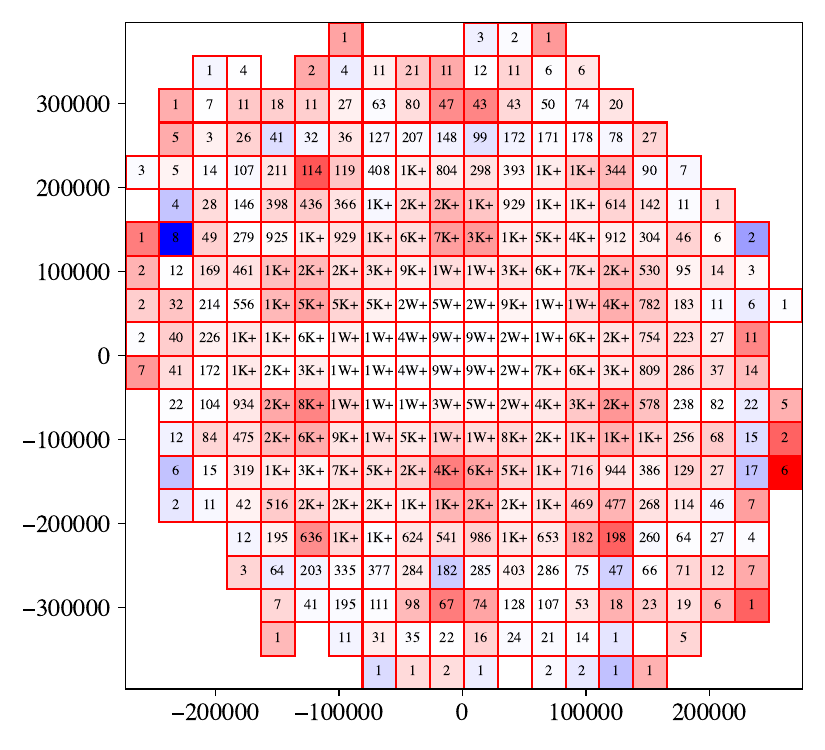}
        \caption{w/ RLE - w/ RLE'}
        \label{fig:sub6}
    \end{subfigure}
    \vspace{-0.3cm}
    \caption{Comparison between different modules.}
    \label{fig:comparison} 
    \vspace{-0.3cm}
\end{figure*}

Taking the experimental area in California as an example, as depicted in Figure \ref{fig:comparison}, we aimed to further assess the effectiveness of various components of our model. We achieved this by analyzing the relative spatial distribution of the differences between the predicted and actual commuting flows and comparing different models or variants of the model through subtraction.

We started by organizing the comprehensive dataset of commuting flows based on their relative locations, subsequently dividing these into distinct cells. In this study, the data was segmented into cells measuring 20 by 20 meters.
Within each cell, we calculated the average residual value. This involved first identifying the difference between the actual and predicted values for two models being compared, such as model A and model B, within the spatial extent of that cell. We then subtracted the residual value of model A in that cell from model B's value and computed their average. This residual value difference is represented by color. Red indicates a positive residual value, meaning model A performs better than model B within that relative spatial range. Conversely, blue indicates a negative residual value, where model A performs worse than model B. The deeper the color, the greater the absolute value of the performance difference is. 

Figure \ref{fig:sub1} essentially illustrates the performance improvement brought by the combined effect of the transformer and RLE modules. Figure \ref{fig:sub2} underscores the contribution of the transformer to performance improvement, while Figure \ref{fig:sub3} focuses on the performance enhancement attributed to the RLE alone. It is evident that our proposed modules enhance performance across almost all relative spatial ranges, with a few exceptions in cells at great relative distances, which have very few flows, often in single digits, and can be considered outliers.

Moreover, we further explore the performance differences between two types of relative location encoders. Figure \ref{fig:sub4} shows the residual map of true versus predicted values using RLE’, where darker green cells indicate poorer performance in predicting flows from their origins to destinations within that spatial range, with lighter green cells indicating smaller differences. The areas of deeper blue are roughly distributed at the vertices of a hexagon, which is an artifact created due to the use of 3 base vectors oriented $2\pi/3$ apart from each other. 
Therefore, to enhance performance and correct this artifact, we experimented with introducing an additional set of base vectors rotated by a certain angle, constructing the RLE. The residual map, as shown in Figure \ref{fig:sub5}, visibly eliminates the hexagonal pattern. Figure \ref{fig:sub6}, depicting the difference between RLE and RLE’, further validates our hypothesis.


\begin{figure}[t]
  \centering
  \begin{subfigure}[b]{0.5\columnwidth}
    \includegraphics[width=\textwidth]{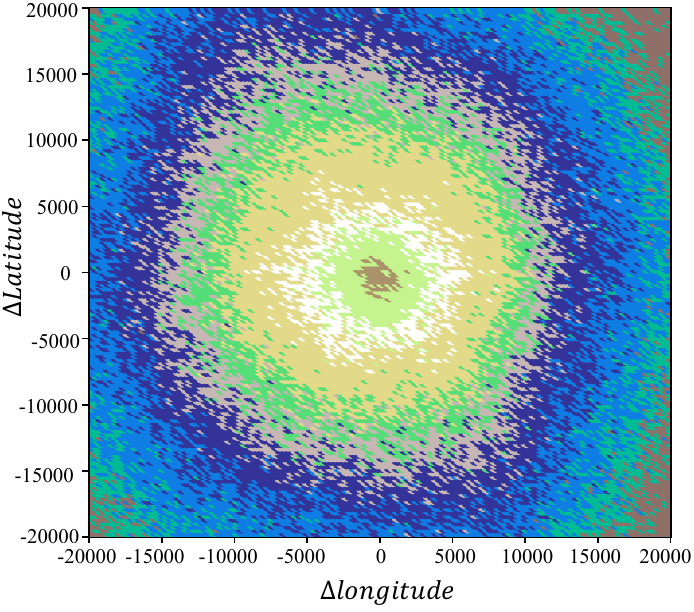}
    \caption{Catesian Coordinate System}
    \vspace{-0.3cm}
    \label{fig:rle_clustering:a}
  \end{subfigure}%
  \begin{subfigure}[b]{0.5\columnwidth}
    \includegraphics[width=\textwidth]{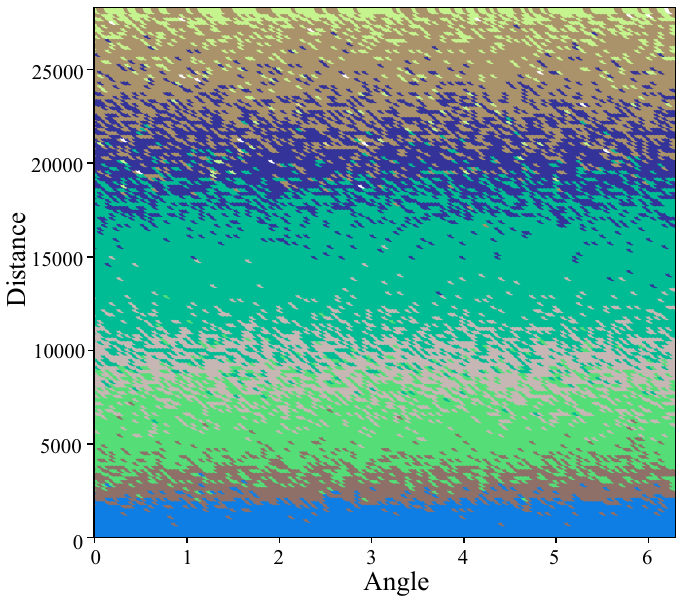}
    \caption{Polar Coordinate System}
    \vspace{-0.3cm}
    \label{fig:rle_clustering:b}
  \end{subfigure}
  \caption{Location embedding clustering of the relative location encoder from the trained \model{}.}
  \vspace{-0.3cm}
  \label{fig:rle_clustering}
\end{figure}

\subsection{Explainability Analysis}

\begin{figure}[t]
  \centering
  \begin{subfigure}[b]{0.5\columnwidth}
  \centering
    \includegraphics[width=\textwidth]{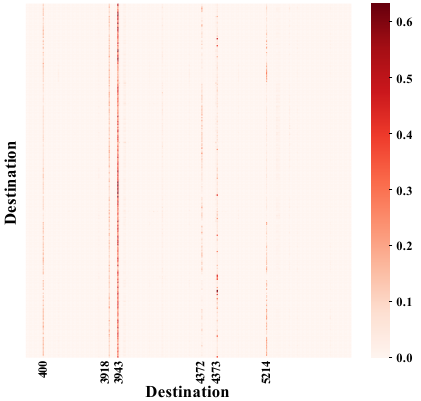}
    \caption{Flow-to-flow attention map}
    \vspace{-0.3cm}
    \label{fig:attmap}
  \end{subfigure}%
  \begin{subfigure}[b]{0.5\columnwidth}
  \centering
    \includegraphics[width=\textwidth]{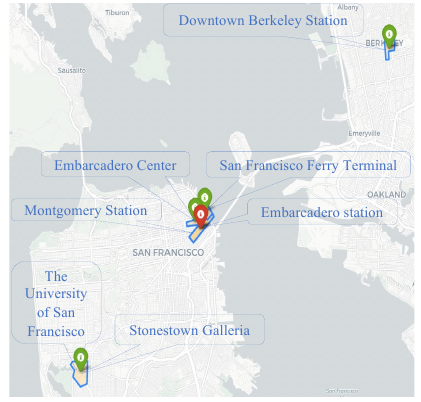}
    \caption{Map visualization}
    \vspace{-0.3cm}
    \label{fig:sanfan_eg}
  \end{subfigure}
  \caption{Case study for flow prediction of one origin region in San Francisco.}
  \vspace{-0.3cm}
  \label{fig:casestudy}
\end{figure}

Grasping the rationale behind a model's prediction is essential for interpreting outcomes, elucidating disparities among models, and gauging our comprehension of the phenomenon being studied.

Our model is elucidated through two primary aspects: Firstly, we investigate the patterns learned by our specially designed relative location encoder, as evidenced by the clustering effectiveness of the generated embeddings. Secondly, we delve into the interpretation of the significance encapsulated within the flow-to-flow attention mechanism.

\subsubsection{Location Embedding Clustering}
We extracted the RLE from the trained \model{} to investigate how it functions. Starting with the maximum diameter $lambda_{max}$ of the study area, we determined the range for visualizing coordinates to be between $[- \lambda_{max}, \lambda_{max}]$. Taking California as an example, where $-\lambda_{max} = 20013$, the range of both longitude and latitude in the Cartesian coordinate system, as shown in Figure \ref{fig:rle_clustering}, spans from $[-20013, 20013]$. We then divided this range into 100 equal intervals, creating a grid of 100x100 cells. The coordinate ranges of these 100x100 cells were fed as inputs into the previously trained RLE to generate embeddings for each cell. Subsequently, we applied a hierarchical clustering algorithm to cluster the cells, opting for a 10-category classification and assigning different colors to different categories. To enhance clarity, we also performed visualization in polar coordinates. Figure \ref{fig:rle_clustering:a} demonstrates that the clusters of location embeddings primarily form based on their relative distance. However, there is also a subtle variation in clustering by direction, as evidenced by the clusters not forming perfect concentric rings.
This indicates that the embeddings obtained from the relative location encoder are anisotropy-aware and capable of capturing both direction and distance.

\subsubsection{Flow-to-Flow Attention Map}

To gain deeper insights into what the model learns from the interactions between flows, we created a flow-to-flow attention map. We select an origin region in San Francisco as an example, for which the 256 predicted destinations formed a 256x256 attention matrix. The sum of each row in this matrix equals 1. As depicted in Figure \ref{fig:attmap}, we visualize this matrix where the y-axis represents the destinations, the x-axis denotes the regions influencing the predicted probability of these destinations, and the color intensity signifies the attention values. The darker red points indicate that the x-axis region has a more significant influence on the flow prediction for the y-axis corresponding destination region. Remarkably, the map displays several vertical red lines, suggesting that some regions consistently have a substantial impact on the flow predictions for that origin region.

Naturally, we seek to further investigate what these influential regions correspond to. As shown in Figure ~\ref{fig:sanfan_eg}, we mark these regions on the map: the origin region in a red icon and the significantly influential regions in green icons. It is observed that these regions carry distinct significance, such as the central business district, the train station, downtown Berkeley, and the University of San Francisco. This demonstrates that our model has learned physically meaningful attention patterns, further emphasizing the critical role of flow interactions in predicting commuting flows.

\section{Conclusion}
This paper tackles the challenge of predicting commuting flows, crucial for urban planning and policymaking. Traditional models like the gravity model and radiation model fall short due to complex variables, whereas deep learning models often compromise explainability for performance. We introduce \model{}, a transformer-based approach with flow-to-flow attention for precise flow predictions. It features an anisotropy-aware geo-spatial encoder for detailed flow analysis, processed by a transformer with flow-to-flow attention to understand flow interactions. \model{} excels in both explainability and performance, paving the way for more informed and effective urban development strategies.
\bibliographystyle{ACM-Reference-Format}
\bibliography{sample-base}


\end{document}